\documentclass[conference]{IEEEtran}
\IEEEoverridecommandlockouts
\usepackage{cite}
\usepackage{multirow}
\usepackage{amsmath,amssymb,amsfonts}
\usepackage{algorithmic}
\usepackage{graphicx}
\usepackage{textcomp}
\usepackage{xcolor}
\usepackage{url}
\def\BibTeX{{\rm B\kern-.05em{\sc i\kern-.025em b}\kern-.08em
    T\kern-.1667em\lower.7ex\hbox{E}\kern-.125emX}}
\begin{document}

\title{A Self-Refining Framework for Enhancing ASR Using TTS-Synthesized Data}

\author{
\IEEEauthorblockN{
    Cheng-Kang Chou\textsuperscript{*1,2},
    Chan-Jan Hsu\textsuperscript{*1},
    Ho-Lam Chung\textsuperscript{2},
    Liang-Hsuan Tseng\textsuperscript{2},\\
    Hsi-Chun Cheng\textsuperscript{2},
    Yu-Kuan Fu\textsuperscript{3},
    Kuan Po Huang\textsuperscript{2},
    Hung-Yi Lee\textsuperscript{2}
}
\IEEEauthorblockA{
    \textsuperscript{1}MediaTek Research \quad
    \textsuperscript{2}National Taiwan University \quad
    \textsuperscript{3}Nvidia
}
\thanks{\textsuperscript{*}Equal contribution. This work was conducted by Cheng Kang Chou during his research internship at MediaTek Research, in collaboration with research scientist Chan-Jan Hsu. Correspondence to: {b09705011@ntu.edu.tw}.}
}
\maketitle

\begin{abstract}
We propose a self-refining framework that enhances ASR performance with only unlabeled datasets. The process starts with an existing ASR model generating pseudo-labels on unannotated speech, which are then used to train a high-fidelity text-to-speech (TTS) system. Then, synthesized speech text pairs are bootstrapped into the original ASR system, completing the closed-loop self-improvement cycle. We demonstrated the effectiveness of the framework on Taiwanese Mandarin speech. Leveraging 6,000 hours of unlabeled speech, a moderate amount of text data, and synthetic content from the AI models, we adapt \textit{Whisper-large-v2} into a specialized model, \textit{Twister}. \textit{Twister} reduces error rates by up to 20\% on Mandarin and 50\% on Mandarin-English code-switching benchmarks compared to \textit{Whisper}. Results highlight the framework as a compelling alternative to pseudo-labeling self-distillation approaches and provides a practical pathway for improving ASR performance in low-resource or domain-specific settings.

\end{abstract}

\section{Introduction}

Automatic speech recognition (ASR) has become a cornerstone technology in modern human–computer interaction, powering applications such as voice assistants, real-time transcription, and accessibility tools. As ASR systems continue to evolve, the demand for transcribed speech data has increased substantially. State-of-the-art generalist ASR models leverage large-scale speech datasets, generally involving at least 25,000 hours and scaling up to millions of hours \cite{nvidia2023canary, radford2022whisper, phi4mini2024, zhang2023usm}. 

Even with these advances, the growing spectrum of multilingual and domain-specific applications calls for additional attention to model development. Replicating the success of supervised pair training across diverse use cases remains challenging, primarily due to the limited availability of transcribed speech data. In contrast, textual data is considerably more abundant \cite{zhang2023usm}. Contemporary textual corpora curated for generative language models\cite{touvron2023llama, groeneveld2024olmo}, when converted into spoken form, can yield speech data volumes over 100 million hours, far exceeding current reserves.\footnote{A 15TB corpus roughly equates to 2-3 trillion tokens. Converting to spoken form with a spoken speed of 120 tokens per minute, this corresponds to more than 100 million hours of speech data.} This virtually unlimited data resource presents a compelling opportunity to address the scarcity of speech data, particularly in underrepresented scenarios such as low-resource or tail languages, code-switching contexts, and specialized domains. There is thus a strong incentive to leverage text-to-speech systems (TTS) to generate paired speech for textual data, to leverage these resources in supervised approaches.


While previous research has touched on the use of TTS systems in scenarios such as personalization \cite{zheng2021using, bataev2023text, yang2023text} and domain adaptation \cite{su2024corpus, su2024task}, these applications have largely remained limited in scope until \cite{yang2025enhancing}. The emergence of high-fidelity controllable TTS systems \cite{coqui_tts, mehta2024matcha, du2024cosyvoice, hsu2025breezyvoice} has made it feasible to generate large-scale synthetic speech corpora with realistic prosody and acoustic variability. The high similarity of this synthetic speech to real speech, as reflected by MOS scores, presents new opportunities to extend prior work and enhance the domain adaptability of ASR systems.



In this work, we draw inspiration from recent advancements and propose a generalizable, self-refining framework for ASR systems. Contrary to prior work \cite{yang2025enhancing}, our work focuses on enhancing existing models on the target language without any handcrafted paired data. The process starts with an existing ASR model generating pseudo-labels on unannotated speech, which are then used to train a high-fidelity
text-to-speech (TTS) system, a step that has been covered in prior work \cite{hsu2025breezyvoice}. 
Then, we create synthesized speech text pairs from the TTS model and bootstrap them into the original
ASR system, completing the closed-loop self-improvement cycle. Throughout the process, only unlabeled single modality data is needed, making the framework highly extensible.



To illustrate the effectiveness of our framework, we focus on adapting \textit{Whisper-large-v2} to Mandarin, a homophonic language that adds an additional layer of modeling difficulty. We also include the setting of Mandarin-English mixing scenarios to account for the frequent inclusions of code-switching content in real-world communications in Asian linguistic environments. In our demonstration, only 6,000 hours of unlabeled audio and less than 1GB of text (to generate 10,000 hours of synthetic content) is used to fuel the training process. 

The resulting ASR model, referred to as \textit{Twister}, achieves substantial performance gains compared to its base model \textit{Whisper-large-v2}. \textit{Twister} reduces error rates by up to 20\% on Mandarin and 50\% on Mandarin-English code-
switching benchmarks. Compared to traditional pseudo-labeling self-distillation approaches \cite{radford2022whisper, tseng2024leave}, the inclusion of the TTS model in the self-refinement loop lowers the real speech data required by 10x, while reaching comparable or better performance. Moreover, this process has the added advantage of being further scalable along two dimensions: by incorporating additional textual content to generate more synthetic speech pairs during the TTS generation phase, and by performing the refinement loop iteratively.



Overall, we present a framework that highlights the potential of TTS-generated speech as a practical alternative to real audio for improving ASR models. We envision this approach as a scalable, flexible, and accessible solution, especially for resource-constrained ASR applications. To support further research, we have open-sourced our model and the accompanying synthetic datasets.

\section{Related Work}


\subsection{Non-English ASR}
Despite the strong multilingual performance of recent ASR models, many prior studies have express needs to further improve recognition in low-resource languages using either real \cite{gete2025whispering, kummervold2024whispering, shekoufandeh2025improving, rijal2024whisper, bajo2024efficient, timmel2024fine} or pseudo-labeled data \cite{tseng2024leave, rijal2024whisper}. However, such data for long-tail languages is much more limited compared to English, with fewer domains represented and primarily consisting of evaluation datasets \cite{lu-etal-2023-ntnu}. Beyond the challenge of acquiring large-scale, realistic data, homophonic languages such as Mandarin introduce an added layer of complexity, as pseudo-labeling can easily reinforce incorrect but identically sounding homophones. In contrast, typographically rendered text produced through precise inputs is typically much more accurate. Motivated by these challenges, we adopt Mandarin ASR as a representative case study of the broader lower-resource ASR problem. Through this lens, we investigate the use of synthetic speech as a scalable approach to addressing data scarcity in the development of ASR models.

\subsection{Code-switching ASR}
Code-switching refers to the practice of alternating between two or more languages within a conversation or even a single utterance, and is commonly present in real-world communications in Asian linguistic environments. Traditional ASR models trained on monolingual corpora struggle to generalize across language boundaries, leading to degraded performance in this field. Common decoding errors include misinterpreting code-switching as direct translation or producing phonetic approximations based on the original language. Mitigation efforts of prior work \cite{huang2024zero, yang2024investigating, liao2023zero} include providing multiple language tokens \cite{peng2023prompting}, or using Speech In-Context Learning (SICL) \cite{wang2024can}. We explore synthesizing code-switching data on the utterance level to enhance code-switching capabilities of the model.

\subsection{Realistic TTS}
Achieving realistic TTS relies on two key elements: precise modeling of the textual input and the production of natural-sounding speech. In recent developments, Large Language Models (LLMs) have been increasingly leveraged for text modeling, owing to their sophisticated understanding of semantic content and a degree of inherent knowledge of pronunciation. This comprehensive linguistic competence enables LLMs to more effectively address the sequence-to-sequence challenge of mapping text tokens to corresponding speech tokens. The representation of speech tokens have largely transitioned from unsupervised \cite{defossezhigh, du2024funcodec} to supervised approaches \cite{an2024funaudiollm}, resulting in increased semantic information density and improved alignment with text.  Finally, prior work has highlighted the efficacy of optimal-transport conditional flow matching (OT-CFM) \cite{mehta2024matcha, du2024cosyvoice, hsu2025breezyvoice} in generating speech.

\subsection{ASR with Synthetic Data}

The dual nature of speech production and perception in human cognition was introduced as early as the Speech Chain concept \cite{denes1993speech}, which was later captured and modeled by deep learning methods \cite{tjandra2017listening, tjandra2018machine, nakayama2018speech}. Recent progress in large-scale supervised ASR and TTS systems has further advanced the enhancement of ASR using synthetic speech in real-world audio scenarios \cite{zheng2021using, bataev2023text, yang2023text, su2024corpus, su2024task, yang2025enhancing}. We extend prior work by scaling up both the quantity and quality of data, which not only generalizes the approach to model natural speech in the wild, but also enables direct comparisons with another prominent direction of pseudo-labeling self-distillation ASR enhacement \cite{radford2022whisper, tseng2024leave}.


\section{Methods}

    
    
    
    
    





\begin{figure*}[htbp]
  \centering
  \includegraphics[width=0.9\textwidth]{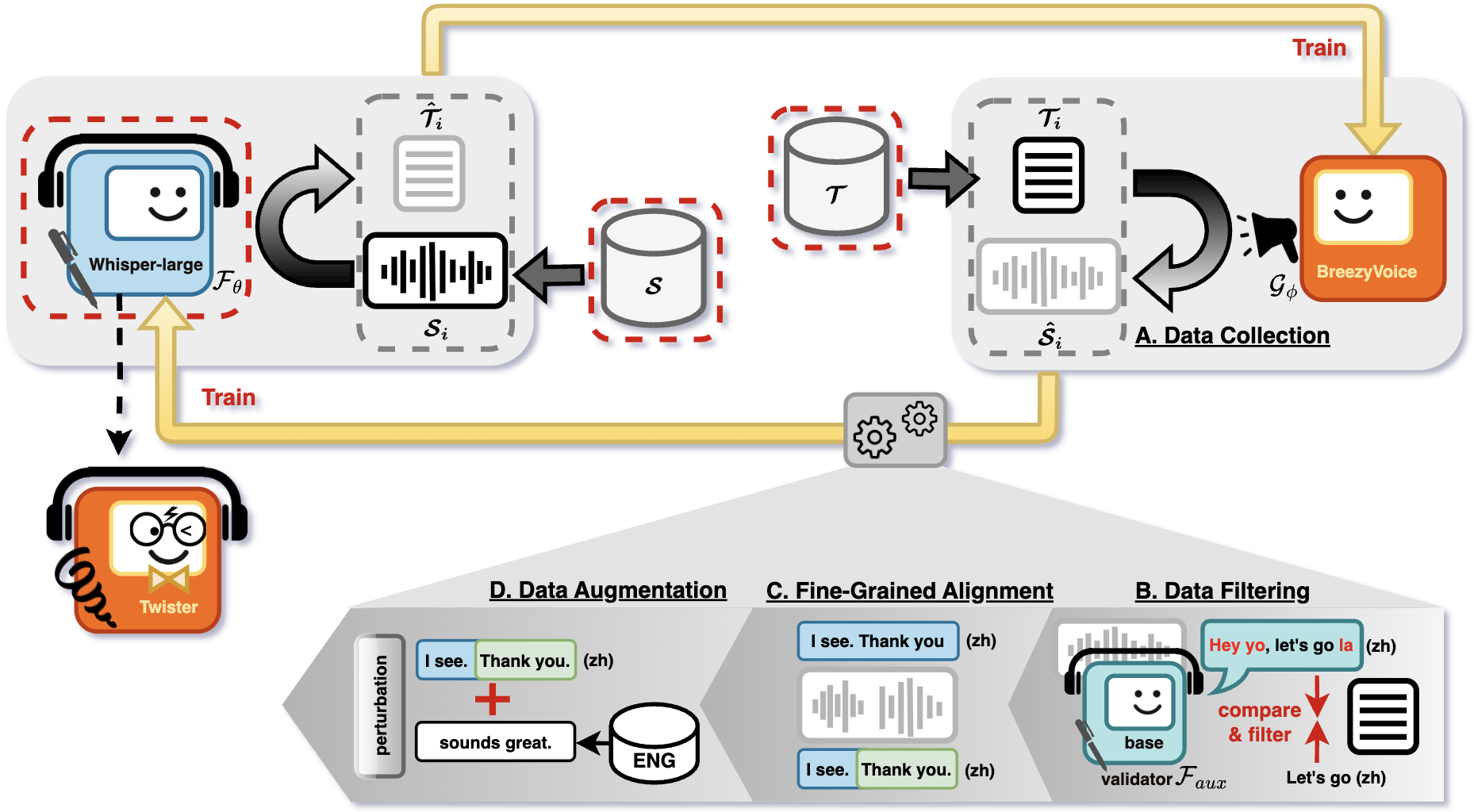}
    \caption{Overview of the self-refining framework. The process begins by generating pseudo-labeled speech–text pairs from a pre-existing ASR model(framed by red dashed lines, ${\mathcal{F}_\theta}$) and a collection of unpaired speech(framed by red dashed lines, $\boldsymbol{\mathcal{S}}$). A TTS model is trained on this pseudo-labeled data and subsequently used to synthesize speech from a large-scale pre-collected text corpus(framed by red dashed line, $\boldsymbol{\mathcal{T}}$). To ensure data quality, filtering and forced alignment are applied. To support long-form transcription and code-switching, utterance concatenation is performed. Additionally, random audio perturbations are introduced to enhance robustness against acoustic variability. The final curated dataset is used to train the target ASR model.}
  \label{fig:enter-label}
\end{figure*}



In our framework, we employ a specialized TTS system to refine an ASR system, where the TTS system is trained on pseudo-labels from the ASR system. This self-refinement process is illustrated in Fig~\ref{fig:enter-label}, which what we need initially are three objects framed by red dashed lines. The required data consists solely of $N$ unpaired speech $\boldsymbol{\mathcal{S}} = \{\mathcal{S}_i\}_{i=1}^N$ for generating pseudo-labels $\boldsymbol{\hat{\mathcal{T}}}= {\hat{\{\mathcal{T}_i\}}}^N_{i=1}$ via the ASR and $M$ unpaired text corpus $\boldsymbol{\mathcal{T}}= \{\mathcal{T}_i\}_{i=1}^M$ for generating paired synthetic speech $\boldsymbol{\hat{\mathcal{S}}} = \{\hat{\mathcal{S}_i}\}_{i=1}^M$ from the TTS. Specifically, ($\boldsymbol{\mathcal{S}}$, $\boldsymbol{\hat{\mathcal{T}}}$) trains the TTS, while ($\boldsymbol{\hat{\mathcal{S}}}$, $\boldsymbol{\mathcal{T}}$) is then processed to train the ASR. 




In light of recent progress in realistic TTS models, we build on existing TTS models and refer to prior work for their training details, focusing our methodology on training the ASR system. Consistent with previous approaches to ASR training, a core part of our modular pipeline is on the curation of data, which includes two phases: \textbf{data collection} (Section~\ref{sec:data_collection}) and \textbf{data filtering} (Section~\ref{sec:data_filtering}). To address the distributional gap between synthetic and real data, we incorporate augmentation techniques including  \textbf{alignment} (Section~\ref{sec:forced_alignment}), \textbf{concatenation and random perturbation} (Section~\ref{sec:data_mixing}). Finally, we bootstrap the original ASR model with the processed semi-supervised pairs to enhance its performance. 


\subsection{Data Collection}
\label{sec:data_collection}
We start our pipeline with collected speech $\boldsymbol{\mathcal{S}}$ and a ASR model $\mathcal{F}_\theta$ to generate pseudo-labels $\boldsymbol{\hat{\mathcal{T}}}$, the data pairs are collected to be  $\mathcal{D}_{pseudo}:=\{(\mathcal{S}_i, \hat{\mathcal{T}}_i)\}_{i=1}^N$, which served as the training data for our TTS model $\mathcal{G}_\phi$.
The trained TTS model $\mathcal{G}_\phi$ is then served as a speech generator based on the collected corpus $\boldsymbol{{\mathcal{T}}}$, based on the size of the corpus, we can virtually unlimited speech-text pairs with the dataset described as:
$$\mathcal{D}_{synthesized}:=\{(\hat{\mathcal{S}_i}, {\mathcal{T}}_i)\}_{i=1}^M$$
where $\hat{\mathcal{S}_i} := \mathcal{G}_\phi({\mathcal{T}}_i)$.

\subsection{Data filtering: }
\label{sec:data_filtering}
In generative models such as text-to-speech (TTS), hallucination can occasionally occur, where the synthesized speech includes content not present in the given prompt. Common manifestations of such errors include the insertion of unrelated lexical material and the generation of unintelligible speech.
This mismatch between the generated audio and the intended text can propagate to automatic speech recognition (ASR) models trained on such data, leading to systematic omissions of certain patterns of words and phrases during inference.


To address these issues, we incorporate a pre-filtering stage guided by a lightweight auxiliary model, $\mathcal{F}_{aux}$, or called \textbf{validator}, to filter out low-quality data pairs from $\mathcal{D}_{synthesized}$. The quality of each pair is estimated by measuring the similarity between given corpus $\mathcal{T}_i$ and transcriptions generated by $\mathcal{F}$, i.e., $\mathcal{T}^{aux}_i:=\mathcal{F}_{aux}(\hat{\mathcal{S}_i})$. 
We define a metric with phoneme error rate (PER) of each sample as $\delta_i := \text{PER}(\mathcal{T}_i, \mathcal{T}^{aux}_i)$ with maximum threshold $\alpha$. 
Data points that deviate significantly from the original text, indicated by the PER exceeding the threshold (with $\delta_i \geq \alpha$), are removed from the dataset. The remaining dataset is defined as follows:

$$\mathcal{D}_{filtered} := \{(\hat{\mathcal{S}_i}, \mathcal{T}_i) \mid \delta_i\leq \alpha\}$$


The deliberate choice to compare phonemicized versions of the text allows us to discount orthographic variations, such as homophones, which are not indicative of hallucination errors. If the validator's transcription significantly deviates from the original input text. 

\subsection{Fine-grained Alignment}
\label{sec:forced_alignment}


Although the TTS audio is synthesized from a predefined text corpus, fine-grained alignment between the speech and text with respect to prosodic boundaries is not revealed during the generation process. The explicit fine-grained alignment is important for two reasons. First, timecode supervision is indispensable for models that handle long-range audio using a divide-and-conquer approach. For example, \textit{Whisper} has a maximum speech context length of 30 seconds. Audio longer than this must be split into smaller segments, with ASR applied to each segment individually before the transcripts are stitched together (More details in Section~\ref{sec:data_mixing}). In this process, precise timecodes are essential to enable seamless merging of segments and ensure optimal overall performance. 
Second, from an application standpoint, ASR systems are widely used to generate subtitles, and accurate phrase-level segmentation is crucial to make subtitles easy for viewers to follow.



To mitigate this issue, we employ Montreal Forced Aligner \cite{mcauliffe2017montreal} to conduct forced alignment, resegmenting the transcriptions to fine-grained snippets based on the acoustic characteristics of the synthesized speech, rather than relying solely on the per-instance boundary of the original speech-text pair. This process yields time-aligned speech segments with durations of approximately 3 to 5 seconds. Consequently, the model is exposed to segmentation cues that closely correspond to the underlying speech signal, thereby enhancing its ability to generalize to authentic prosodic variations encountered in real-world scenarios.



\subsection{Data Augmentation}
\label{sec:data_mixing}





\subsubsection{Preserving Long-form ASR Capabilities}

For ASR model $\mathcal{F}_\theta$ that accepts audio inputs up to $L_{\text{max}}$ seconds per inference window (e.g., 30 seconds for \textit{Whisper}), clips exceeding this duration are classified as \emph{long-form}. Long-form audio modeling requires segmenting the input into shorter chunks to satisfy the model’s input constraints. Training the model solely on short utterances (typically 5 to 25 seconds) creates a distributional mismatch between the training data and the $L_{\text{max}}$-second segments encountered during long-form transcription and degrades performance. 
To address this challenge, we augment $L_{max}$-second audios from existing samples to simulate chunked long-form audio.

To obtain long-form audio from short-form ones, we recursively append utterances to an existing clip until the total length slightly exceeds \(L_{\max}\), and cutoff the audio at the \(L_{\max}\) second mark. However, pairing this with the transcription corresponding to the first \(L_{\max}\) seconds (denoted as $\mathcal{T}_{\leq L_{\max}}$) is suboptimal, as this introduces segmentation discontinuities that destabilize chunk merging. Instead, we backtrack to the nearest prosodic boundary \(L_{bound}\) within the \(L_{\max}\) time constraint, identified through forced alignment (cf. Section~\ref{sec:forced_alignment}).
The transcript is then truncated at \(L_{bound}\), omitting any text corresponding to speech in \((L_{bound},L_{\max}]\). A special tag is appended to the end of the text sequence to indicate continuation. During inference, the subsequent chunk begins from the last detected prosodic boundary.
\subsubsection{Enhancing Code-switching ASR Capabilities}

Given the limited availability of open-source datasets with code-switching, synthesizing code-switching data becomes crucial for developing robust multilingual models. To this end, we simulate code-switching scenarios by concatenating utterances from different languages up to a maximum duration of $L_{\text{max}}$ seconds. While this approach primarily reflects sentence-level code-switching, it provides a reasonable approximation of word-level code-switching observed in real-world data.

\subsubsection{Random Audio Perturbations}
Finally, we apply random audio perturbations such as background noise injection and temporal blurring to improve the model’s robustness to acoustic variability.
\subsection{Data Mixing}

While not a distinct processing step, we ensured that all types of data are thoroughly represented in the final corpora. The resulting datasets exhibit multi-faceted diversity, comprising a balanced mix of long and short audio segments across English, Mandarin, and code-switching utterances.

\section{Experiments}

We demonstrate the effectiveness of the framework by selecting Taiwanese Mandarin as our adaptation target.  
The refined ASR model is named as \textit{Twister} (TTS-enhanced Whisper), which supports English, Mandarin, and bilingual code-switching settings.

\subsection{Model}
We utilize \textit{Whisper-large-v2} as our ASR model, denoted as $\mathcal{F}_\theta$, due to its reasonable performance on the target languages. While this model alone suffices for the backbone of our framework, recent advancements in developing a corresponding $\mathcal{G}_\phi$ have yielded encouraging outcomes.  Namely, \textit{BreezyVoice},  a Taiwanese Mandarin TTS system based on the \textit{CozyVoice} architecture, demonstrates state-of-the-art, highly realistic speech synthesis. Given its strong performance, we adopt \textit{BreezyVoice} directly as $\mathcal{G}_\phi$, and refer readers to the original work for details regarding its training procedure of the TTS.
We use \textit{Whisper-base} as the validator to prefilter synthesized data pairs.



\subsection{Data}


Our raw datasets $\{(T_i, \hat{S}_i)\}_{i=1}^N$ include three types of speech sources: Mandarin, English, and Mandarin-English code-switching. 
For Mandarin, we curated textual content from ODC-By licensed FineWeb2 \cite{penedo2024fineweb-2}. Subsequently, we synthesized a large-scale speech corpus of approximately 10,000 hours using \textit{BreezyVoice} (Section~\ref{sec:data_collection}), corresponding to $\mathcal{D}_{synthesized}$. To ensure acoustic diversity and speaker variation, we synthesized speech using voice samples from over 200 Mandarin speakers. To meet the demands of English and code-switching data, we incorporated open-source ASR datasets \textbf{CommonVoice} \cite{commonvoice:2020} and \textbf{NTUML2021} \cite{yang2024investigating}, which are distributed under permissive licenses.
The distribution and volume of these raw datasets are summarized in Table~\ref{tab:datasize}. 
These datasets serve as the basis for further data augmentation.

\begin{table}[htbp]
\centering
\caption{Data Distribution of Collected Raw Corpus}
\begin{tabular}{|l|c|c|c|}
\hline
\textbf{Dataset Name} & \textbf{Type} & \textbf{Language} & \textbf{Total Hours} \\
\hline
ODC Synth & Synth. & Mandarin & 10,000 \\
CommonVoice17-EN & Real & {English} &  {1,738} \\
NTUML2021 & Real & Code-switching & 11 \\
\hline
\end{tabular}
\label{tab:datasize}
\end{table}

We set the data filtering threshold alpha to 0.6, using the validator model \textit{Whisper-base}, resulting in a filtered Mandarin speech dataset with 4,000 hours of audio remaining. The audios are subsequently aligned to obtain fine-grained timestamps. At this stage, the dataset is fully prepared for augmentation.

The augmentation process aims to construct two distinct subsets: a Mandarin set and a mixed (code-switching) set. The Mandarin set is formed exclusively from concatenating Mandarin clips sampled from \textbf{ODC-Synth}, and some short-form English clips are preserved for replay purpose. For the mixed set, we expand the limited natural code-switching corpus \textbf{NTUML2021} by creating artificial code-switching samples through random mix-and-match combinations of English clips from \textbf{CommonVoice17} and Mandarin clips from \textbf{ODC-Synth}. Table~\ref{tab:datasizefinal} summarizes the data quantities of the final dataset, which comprises a mix of long and short audio segments in English, Mandarin, with English included minimally solely to mitigate forgetting.  During training, data points are sampled uniformly from all instances across all datasets.

\begin{table}[htbp]
\centering
\caption{{Data Distribution of Final Training Corpus}}
\begin{tabular}{|l|c|c|c|c|}
\hline
\textbf{Dataset Name} & \textbf{Type} & \textbf{Length} & \textbf{Language} & \textbf{Total Hours} \\
\hline
\textbf{}Mandarin Long & Synth. & Long & Zh &  4,000 \\
Mandarin Short & Synth. & Short & Zh &  70 \\
English & Real & Short & En &  10\\ 
Code-switching & Real+Synth & Long & C.S. & 1,715 \\
\hline
\end{tabular}
\label{tab:datasizefinal}
\end{table}

\subsection{Training Details}
To enable unified processing of Mandarin, English, and code-switching utterances, we initialized a shared language embedding by taking the element-wise average of the language-token embeddings for \textlangle\textbar zh\textbar\textrangle and \textlangle\textbar en\textbar\textrangle. This strategy provides a language-neutral initialization that is universal across intended use cases. This approach is inspired by the zero-shot inference results of the base model, \textit{Whisper-large-v2}. As shown in Table~\ref{tab:mer_whisper}, the mixed-language embedding achieves comparable performance on monolingual Mandarin and English benchmarks, and even improves accuracy on code-switching tasks, despite not being explicitly trained for this configuration.


We train \textit{Whisper-large-v2} on the described dataset, totaling 10000 steps with a batch size of 256. The learning rate was set
to $2 \times 10^{-5}$. Training was conducted on 8 NVIDIA H100 GPUs. 

\begin{table}[htbp]
    \centering
    \caption[MER of \textit{Whisper-large-v2} across language tags]%
    {MER of \textit{Whisper-large-v2} across datasets using different language tags. Mixed embedding performs best at code-switching scenarios while retaining performance on monolingual sets.}
    \label{tab:mer_lang}

    \begin{tabular}{|l|r|r|r|}
        \hline
        \textbf{Dataset\textbackslash Setting} & \textbf{Mixed} & \textbf{ZH} & \textbf{EN} \\ \hline
        ASCEND-EN       & 29.63 & 101.54 & 27.20 \\ 
        ASCEND-ZH       & 17.65 &  13.75 & 78.50 \\ 
        ASCEND-Mixed$^{\ast}$ & 21.90 & 22.69 & 80.61 \\ 
        CommonVoice16-zh-TW & 11.94 &  9.02 & 42.08 \\ 
        CSZS-zh-en$^{\ast}$  & 26.25 & 44.28 & 54.27 \\ 
        ML-2021-long$^{\ast}$ &  6.82 &  6.13 & 94.76 \\ \hline
        \multicolumn{4}{l}{\footnotesize $^{\ast}$\,Code-switching datasets.}
    \end{tabular}
    \label{tab:mer_whisper}
\end{table}

\subsection{Evaluation Data}
For performance evaluation, we use different datasets to cover Chinese-Mandarin, Taiwanese-Mandarin, English, code-switching, short-form, long-form datasets.
\subsubsection{ASCEND}
ASCEND \cite{lovenia2022ascend} is a spontaneous, multi-turn conversational dialogue corpus featuring Chinese-English code-switching, collected in Hong Kong. To further analyze the impact of language on performance, we partitioned the dataset into three subsets: EN (English-only), ZH (Mandarin-only), and Mixed (Mandarin-English code-switching). 

\subsubsection{CommonVoice16}
CommonVoice16-zh-TW \cite{commonvoice:2020} is a dataset for Taiwanese-Mandarin, which is a subset of CommonVoice 16-1 version, which is publicly available, serving as an evaluation set for assessing short-form Mandarin ASR performance.

\subsubsection{CSZS-zh-en}
CSZS-zh-en \cite{huang2024zero} is a dataset that contains code-switching data adopted from the Amazon Polly text-to-speech system to synthesize utterances.

\subsubsection{ML-lecture-2021-long}
ML-Lecture-2021-Long \cite{yang2024investigating} is a testing dataset comprising approximately 5 hours of recordings derived from the "Machine Learning" course at National Taiwan University, which is publicly available. It features code-switching utterances with a predominance of Taiwan-Mandarin, serving as an in-domain evaluation set for assessing long-form code-switching ASR performance.

\subsubsection{FormosaSpeech} 

FormosaSpeech \cite{lu-etal-2023-ntnu} is a multi-speaker evaluation benchmark for Taiwanese Mandarin, comprising both news and text reading materials, and featuring monologues as well as dialogues.

\subsubsection{Formosa-Suite}
The Formosa Suite represents our own in-domain in-house Taiwanese Mandarin speech corpora designed to evaluate long-form ASR performance. It comprises four subsets covering various subjects: \textbf{Formosa-Go} (tourism and location narratives), \textbf{Formosa-Show} (talk shows and stand-up comedy), \textbf{Formosa-Course} (online-course lectures across academic disciplines), and \textbf{Formosa-General} (a broad mix of topics including technology, lifestyle, and food, etc.). Each subset contains ~3-minute clips and ranges from 5 to 10 hours in total test duration, collectively covering a diverse range of speaking styles, domains, and speaker conditions.

\begin{table*}[htbp]
\centering
\caption{MER of Different Taiwan Mandarin ASR systems. Best performing results are marked in \textbf{bold}, and relative word error rate reduction compared to WLV2-auto are marked in brackets. Columns with an asterisk denote code-switching datasets.
 }
\label{tab:main_results}
\small
\begin{tabular}{|l|c|c|c|c|c|}
\hline
\textbf{Dataset\textbackslash Model} & \textbf{WLV2-Oracle}$\downarrow$ & \textbf{WLV2-Auto}$\downarrow$  
 & \textbf{WLV3-auto}$\downarrow$ & \textbf{COOL-Whisper}$\downarrow$ & \textbf{Twister (Ours)}$\downarrow$ \\
\hline
\multicolumn{6}{|c|}{\textbf{Short Audio Datasets}} \\
\hline
ASCEND-OVERALL* \cite{lovenia2022ascend}              & 21.14 (AUTO) & 21.14 & 23.22 & 19.71 & \textbf{17.74 }(-16.08\%) \\
- ASCEND-EN               & 27.20 (EN)   & 27.36 & 27.21 & 29.39 & \textbf{26.64 }(-2.63\%)  \\
- ASCEND-ZH               & \textbf{13.75} (ZH)   & 17.49 & 17.41 & 18.90 & 16.04 (-8.29\%)  \\
- ASCEND-MIX*              & 21.01 (AUTO)& 21.01 & 25.13 & 17.34 & \textbf{16.38} (-22.01\%)  \\

CommonVoice16-zh-TW \cite{commonvoice:2020}                    &  9.02 (ZH)   &  9.84 &  8.95 & 11.86 &  \textbf{7.97 }(-19\%)  \\
CSZS-zh-en* \cite{huang2024zero} & 29.49 (AUTO) & 29.49 & 26.43 & 20.90 & \textbf{13.01} (-55.88\%) \\
\hline
\multicolumn{6}{|c|}{\textbf{Long Audio Datasets}} \\
\hline
ML-lecture-2021-long* \cite{yang2024investigating} &  6.13 (ZH)   &  6.13 & 6.41 &  6.37 &  \textbf{4.98} (-18.76\%)           \\
Formosa-Go              & 15.03 (ZH)   & 15.03 & 14.90 & 16.83 & \textbf{13.61} (-9.44\%)           \\
Formosa-Show            & 29.18 (ZH)   & 29.18 & 27.80 & 29.78 & \textbf{27.58 }(-5.48\%)        \\
Formosa-Course          &  9.50 (ZH)   &  \textbf{9.50} & 9.67 & 11.12 &  9.94 (+0.44\%)           \\
Formosa-General         & 11.45 (ZH) & 11.45 & 11.46 & 13.33 & \textbf{11.37} (-0.69\%)           \\
FormosaSpeech \cite{lu-etal-2023-ntnu} & 22.34 (ZH) & 22.34 & 21.22 & 26.71 & \textbf{22.09} (-1.12\%)            \\
\hline 

\end{tabular}
\end{table*}

\subsection{Evaluation Metrics}

We adopt Mixed Error Rate (MER) as evaluation metric to assess model performance on code-switching speech recognition tasks. MER computes character error rate (CER) for Mandarin segments and word error rate (WER) for English segments, thereby aligning with the natural granularity of individual languages.




\section{Results}

To evaluate the effectiveness of our proposed model, \textit{Twister}, we conduct a comprehensive comparison against a suite of baselines built upon the \textit{Whisper} architecture. The primary baseline is \textit{Whisper-large-v2}, the ASR model prior to self-refinement. 


In addition, we include two models that also claims improvement on Taiwanese Mandarin. \textit{Whisper-large-v3} and \textit{COOL-Whisper}. Coincidentally, both models leverage pseudo-labeling techniques, followed by distillation approaches using the paired data.
\textit{Whisper-large-v3} is an upgraded version of \textit{Whisper-large-v2}, trained on 1 million hours of high-quality speech-text pairs and an additional 4 million hours of pseudo-labeled audio. Using a 4.4\% Mandarin composition in the \textit{Whisper-v1} training data as a reference, we estimate that \textit{Whisper-large-v3} was exposed to approximately 220,000 hours of Mandarin data throughout training. 
\textit{COOL-Whisper} is a lightweight model comparable in size to \textit{Whisper-medium}, with approximately half the number of parameters of \textit{Whisper-large-v2}. It was trained using k2d \cite{tseng2024leave} on a dataset comprising 60,000 hours of Taiwanese Mandarin course materials. Given the inherent code-switching present in these recordings, \textit{COOL-Whisper} provides a strong baseline for evaluation on code-switching benchmarks. We summarize evaluation results in Table~\ref{tab:main_results}. The column \textit{WLV2-Auto} is the standard inference scheme of \textit{Whisper-large-v2} where the language tag is automatically decided by the model, whereas \textit{WLV2-Oracle} denotes the best-case performance among three different inference configurations (automatic detection, forced Mandarin token, and forced English token).



\subsection{Comparisons with \textit{Whisper-large-v2}}

Compared to the original \textit{Whisper-large-v2} model, \textit{Twister} demonstrates significant performance improvements on both Chinese and code-switching datasets. Notably, on the CSZS benchmark, it achieves a substantial WERR of 55.88\%, prior methods that rely on more elaborate evaluation schemes \cite{yang2024investigating}.
In benchmarks where automatic language detection underperforms compared to settings where the language is provided (i.e., WLV2-Auto $>>$ WLV2-Oracle), \textit{Twister} demonstrates moderate to substantial improvements. Specifically, it achieves relative WERRs of 8.29\% and 19\% on ASCEND-ZH and CommonVoice16-zh-tw, respectively. This suggests that our mixed embedding approach effectively mitigates language detection errors, particularly in cases involving noisy or ambiguous audio inputs. Interestingly, a slight performance gain is also observed on ASCEND-EN, which we attribute to the use of synthetic data simulating Asian-accented English. Finally, we observe consistent gains on long-form audio, indicating that our augmentation method successfully extends short-form data into plausible long-form variants that transfer well to real-world long-form speech. Overall, the performance improvements of Twister compared to Whisper-large-v2 demonstrate that the framework emerges as a positive reinforcement loop for self-refinement. These results encourage future research to explore further scaling in the TTS generation phase or dynamics of iterative refinement.

\subsection{Comparisons with other Models}
Additional baselines included in our comparison are \textit{Whisper-large-v3} and \textit{COOL-Whisper}, both of which utilize large-scale audio datasets and \textit{Whisper-large-v2} in conjunction with a distillation methodology. From Table~\ref{tab:main_results}, it can be observed that \textit{Twister} performance outperforms other approaches in all but one benchmark. Our findings emphasize the critical role of content in distilling speech-text pairs for ASR, aligning with prior TTS-based approaches \cite{yang2023text}. Moreover, our method is notably more data-efficient: whereas previous methods utilize over 60,000 hours of raw speech data, Twister achieves superior results with at least 10 times less. These findings demonstrate that integrating TTS into the framework effectively decreases the dependence on real speech data, addressing a longstanding bottleneck in low-resource speech modeling.

\section{Conclusion}
In this work, we propose a self-refining framework that enhances ASR performance with only unlabeled datasets. We outline a methodology that leverages high-quality TTS-synthesized audio to bootstrap and enhance the performance of the original ASR system. Results show that our framework is effective in Mandarin and code-switching scenarios for both long-form and short-form audio, with performance gains up to 55.88\% compared to the original model \textit{Whisper-large-v2}. Compared to pseudo-labeling self-distillation approaches employed by \textit{Whisper-large-v3} and \textit{Cool-whisper}, our method achieves comparable or superior performance with significantly higher data efficiency. Overall, our framework addresses the limitations posed by real data scarcity and offers a generalizable solution for adapting ASR systems to low-resource and underrepresented speech domains.

\section*{Acknowledgments}
We thank NVIDIA for providing access to the Taipei-1 supercomputer.
\bibliographystyle{IEEEbib}
\bibliography{refs}



 


\end{document}